# Can large language models unlock discrete data in ophthalmic diagnostic reports?

**Umair A. Zaidi[1]; An-Lun Wu, MD[2]; Wei-Chun Lin, MD, PhD[2]; Thomas S. Hwang, MD[2]; Michelle R. Hribar, PhD[2,3]**

[1]Division of Informatics, Clinical Epidemiology and Translational Data Science, Oregon Health & Science University, Portland, OR
[2]Casey Eye Institute, Department of Ophthalmology, Oregon Health & Science University, Portland, OR
[3]Department of Ophthalmology and Visual Sciences, University of Illinois Chicago, Chicago, IL
*Correspondence: Michelle R. Hribar, PhD, 1747 W Roosevelt Rd, Rm. 247, Chicago, IL 60608*

Presented at the Association for Research in Vision and Ophthalmology (ARVO) Annual Meeting, Denver, Colorado, May 4, 2026.

## Abstract

**Objective:** To assess the accuracy and efficiency of a large language model (LLM) using two prompt strategies for extracting structured data from ophthalmic diagnostic reports in portable document format (PDF).

**Design:** Evaluation of Technology and Software

**Subjects, Participants, and/or Controls:** Twenty deidentified diagnostic reports comprising visual field and three optical coherence tomography (OCT) report types.

**Methods, Intervention, or Testing:** Structured data from twenty de-identified diagnostic reports (Visual Field, OCT Glaucoma Overview, OCT retinal nerve fiber layer (RNFL) Single Exam, OCT Thickness Map; n = 5 each) were extracted using two different GPT-4o-assisted pipelines and were evaluated against a consensus ground truth from manual abstraction. The first pipeline, Schema-Constrained, used Structured Output mode with a predefined JSON Schema and minimal prompt to return schema-compliant JSON directly. The second pipeline, Prompt-Only, extracted text according to a detailed instruction prompt producing a key–value string that was converted to JSON using Python.

**Main Outcome Measures:** Value accuracy, formatting accuracy, and extraction time.

**Results:** Schema-Constrained achieved 100.00% value accuracy for Visual Field and RNFL Single Exam, 97.45% ± 5.69% for Glaucoma Overview, and 98.00% ± 4.47% for Thickness Map; Prompt-Only achieved 100.00% across all four report types. Formatting accuracy was 100.00% for Schema-Constrained across all report types, and 100.00% for Prompt-Only except RNFL Single Exam (90.14% ± 5.98%). Mean per-report extraction time was 56.51 s for manual review versus 5.04 s (Schema-Constrained) and 4.70 s (Prompt-Only), an ~92% reduction.

**Conclusions:** In this small proof-of-concept dataset, general-purpose LLM-assisted pipelines extracted structured data from ophthalmic diagnostic PDFs with high accuracy and substantially reduced processing time. Under the tested configuration, Prompt-Only achieved the highest value accuracy, while Schema-Constrained produced schema-compliant output with 100% formatting accuracy. These complementary strengths support further evaluation of hybrid, validation-aware workflows for research and clinical data abstraction.

## Introduction

Ophthalmic diagnostics such as automated perimetry (visual fields) and optical coherence tomography (OCT) provide quantitative measures that are integral to clinical decision-making and research. In glaucoma, functional loss on visual field testing and structural change in the retinal nerve fiber layer (RNFL), optic nerve head, and macula measured by OCT are used together for diagnosis, risk stratification, treatment targets, and longitudinal monitoring.[1] In retinal disease, OCT is a standard imaging modality for diagnosing and monitoring age-related macular degeneration, diabetic retinopathy, and other maculopathies. OCT-derived biomarkers such as central subfield thickness and nerve fiber layer thickness guide therapy decisions and assessment of the response.[2,3]

Despite their clinical importance, the discrete data within these diagnostics are often contained only in human-readable reports, commonly portable document format (PDFs), that are not readily machine-readable or interoperable across systems. Device vendors frequently use proprietary formats, and report layouts can vary by model and software version, complicating automated extraction and aggregation. The American Academy of Ophthalmology and others have explicitly called for adoption of existing standards (e.g., DICOM) and development of ophthalmology-specific implementation guides to enable seamless data exchange with electronic health records (EHRs) and registries.[4–6] Traditional deterministic approaches (rule-based templates or OCR pipelines) can work when formats are fixed, but they are brittle to layout changes over time and heterogeneous designs between devices that are common in real-world practice.[7,8] As a result, discrete data from these reports are often not directly integrated into EHRs, limiting their potential for clinical decision support, big data analysis, or quality monitoring and improvement.

Recent advances in large language models (LLMs) present an opportunity to bridge this gap by transforming semi-structured or unstructured clinical text into machine-readable, structured outputs. In non-ophthalmic domains, LLM-based systems have achieved strong performance in extracting structured data from clinical notes and radiology reports and even mapping content into standards such as HL7-FHIR.[9–12] Within ophthalmology, natural language processing (NLP) has been highlighted as a promising yet still emerging approach for leveraging the rich unstructured data generated in care, and early studies are beginning to demonstrate feasibility of LLMs for extracting key variables from ophthalmic documentation.[13,14] However, these efforts have predominantly targeted free-text clinical notes, extracting variables such as visual acuity, intraocular pressure, and disease diagnoses from narrative provider documentation.[13] Diagnostic imaging reports represent a distinct document class: they are device-generated, vendor-specific PDFs containing dense numeric fields organized by anatomic sector and laterality within fixed visual layouts, rather than narrative prose. Extracting structured data from these reports requires preserving precise field-to-value associations across heterogeneous templates, a challenge that has received little attention in ophthalmic NLP to date.

In this study, we present a proof-of-concept evaluation of LLM-assisted extraction from ophthalmic diagnostic reports using two different prompting strategies, focusing on both accuracy and consistency of extracted values. By demonstrating the feasibility of this approach, we aim to highlight the potential of LLMs to enhance research, clinical workflows, and interoperability in ophthalmology.

## Methods

### Study Setting

This study was conducted at the Casey Eye Institute, Oregon Health & Science University (OHSU), using a subset of ophthalmic diagnostic reports. The project was designed as a proof-of-concept evaluation of automated data extraction using large language models (LLMs). This study adhered to the tenets of the Declaration of Helsinki and was approved by the Institutional Review Board at OHSU (STUDY00026815) and by the AI Governance Committee.

### Diagnostic Reports

A total of 20 diagnostic reports were sampled from the ophthalmic Picture Archiving and Communication System Continuum (Integrated Ophthalmic Systems, Inc., Woburn, Massachusetts). We sampled 5 each of 4 report types: Single Field Analysis from Humphrey Field Analyzer (Zeiss Oberkochen, Germany) and OCT Glaucoma Overview, OCT Retinal Nerve Fiber Layer Single Exam, and OCT Thickness Map generated by Spectralis OCT (Heidelberg, Germany). We included only reports in PDF format with machine-readable embedded texts to enable direct text extraction. Reports consisting solely of scanned images were excluded, as image-based parsing was outside the scope of this study. All reports were de-identified prior to analysis by removing the 18 HIPAA identifiers, including patient name, medical record number, date of birth, and exam date, from both the PDF metadata and the visible report fields. De-identification was verified by a second reviewer before the reports entered the extraction pipeline. The specific clinical data fields targeted for extraction are listed in Table 1.

**Table 1.** Clinical data fields targeted for extraction. Abbreviations: OS = Oculus Sinister; OD = Oculus Dexter; GHT = Glaucoma Hemifield Test; VFI = Visual Field Index; MD = Mean Deviation; PSD = Pattern Standard Deviation; OCT = Optical Coherence Tomography; BMO = Bruch's Membrane Opening; MRW = Minimum Rim Width; RNFL = Retinal Nerve Fiber Layer; RNFLT = Retinal Nerve Fiber Layer Thickness; HAAT = Horizontal Angle Asymmetry Test; directional sectors: S = Superior, N = Nasal, T = Temporal, I = Inferior; compound sectors: NS = Nasal-Superior, TS = Temporal-Superior, NI = Nasal-Inferior, TI = Temporal-Inferior; G = Global.

| Report Type | Field Name |
|---|---|
| Visual Field | Eye OS/OD |
| | Pupil Diameter |
| | Test Type, Strategy, Fovea, Fixation Monitor, Fixation Target |
| | Fixation Losses, False Positive Errors, False Negative Errors |
| | GHT, VFI, MD, PSD |
| | Threshold Value Points |
| OCT Glaucoma Overview | Eye: OS/OD |
| | Circle Diameter, BMO Area |
| | MRW Values (NS, TS, N, G, T, NI, TI), MRW Classification |
| | RNFLT Values (NS, TS, N, G, T, TI, NI), RNFLT Classification |
| | HAAT (Superior, Inferior) |
| OCT RNFL Single Exam | Eye: OS/OD |
| | Circle Diameter, BMO Area |
| | RNFL Thickness by Sector 1 (S, N, T, I) |
| | RNFL Thickness by Sector 2 (NS, TS, N, G, T, TI, NI) |
| | Classification |
| OCT Thickness Map | Eye: OS/OD |
| | Circle Diameter |

### Manual Data Extraction (Ground Truth)

Manual transcription was performed by a trained research assistant with prior experience in ophthalmic data abstraction. Each report was reviewed in Adobe Acrobat Pro, and data values were transcribed into Microsoft Excel. The primary transcription was independently verified by two additional reviewers (A.W. and U.Z.), who compared each field against the source PDF. Discrepancies identified during verification were resolved by consensus, and the reconciled dataset was designated as the ground truth. The time required for manual extraction was measured using a stopwatch.

### Automated Parsing Workflows

We implemented two different automated parsing workflows using plain text extracted from the PDF reports. Plain text was extracted from each PDF using the pdfplumber and PyMuPDF (fitz) Python libraries, which preserves reading order and numeric tokens from embedded text layers. No OCR was applied, as all reports contained machine-readable text. Page headers, footers, and device watermarks were retained as-is before submission to the model. No normalization of units, minus signs, or laterality codes (OS/OD) was performed prior to LLM input; these conventions were handled by the model according to the prompt or schema specification. Fields not present in a given report were returned as null by the model. The same extracted text string was supplied to both pipelines to ensure a fair comparison. An overview of the end-to-end workflow, from PDF text extraction to LLM parsing and output, is shown in Figure 1.

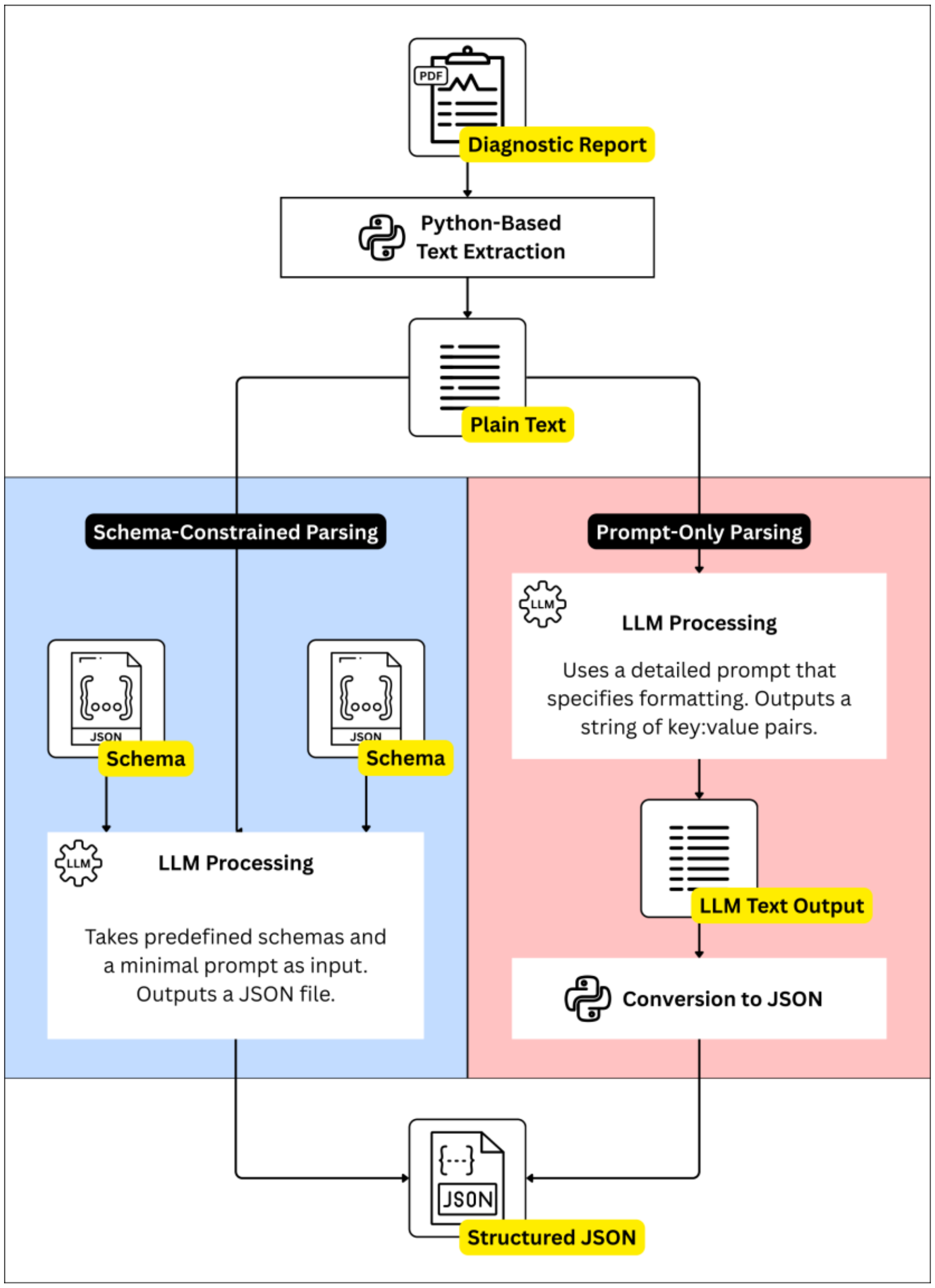

**Figure 1.** Pipeline overview for LLM-assisted extraction. Plain text is extracted from diagnostic PDFs with a Python tool, then processed via two branches: Prompt-Only Parsing (prompted LLM returns a key–value string that is subsequently converted to JSON) and Schema-Constrained Parsing (LLM returns schema-compliant JSON using a predefined schema and minimal prompt).

The extracted text was then processed using two distinct pipelines:

**Schema-Constrained Parsing:** Extracted text was submitted to GPT-4o using Structured Output mode. In this pipeline, the model was provided with a predefined JSON Schema and a minimal prompt (Figure 2). The output was a valid, schema-compliant JSON file, eliminating the need for additional formatting or text-to-JSON conversion (Figure 3).

```
system_prompt = """You are an expert at structured data extraction. You will be given unstructured text
 from an opthalmology report and should convert it into the given structure. If you cannot find a
certain value, leave the field blank."""

RNFL_SCHEMA = {
    "data": {
        ...
        "OD": {
            "type": "object",
            "properties": {
                "Circle Diameter": {"type": "number"},
                "BMO Area": {"type": "number"},
                "Classification": {"type": "string"},
                "Sector Map 1 Superior (S)": {"type": "number", "multipleOf": 1},
                "Sector Map 1 Nasal (N)": {"type": "number", "multipleOf": 1},
                "Sector Map 1 Temporal (T)": {"type": "number", "multipleOf": 1},
                "Sector Map 1 Inferior (I)": {"type": "number", "multipleOf": 1},
                "Sector Map 2 Nasal Superior (NS)": {"type": "number", "multipleOf": 1},
                "Sector Map 2 Temporal Superior (TS)": {"type": "number", "multipleOf": 1},
                "Sector Map 2 Nasal (N)": {"type": "number", "multipleOf": 1},
                "Sector Map 2 Global (G)": {"type": "number", "multipleOf": 1},
                "Sector Map 2 Temporal (T)": {"type": "number", "multipleOf": 1},
                "Sector Map 2 Nasal Inferior (NI)": {"type": "number", "multipleOf": 1},
                "Sector Map 2 Temporal Inferior (TI)": {"type": "number", "multipleOf": 1}
            },
        },
        "OS": {...},
    }
},
```

**Figure 2.** Schema-Constrained configuration. Minimal prompt and excerpt of the predefined JSON Schema used with Structured Output mode to enforce field names, types, and structure.

```
"parsed": {
  "file_name": "redacted_122.pdf",
  "report_type": "RNFL Single Exam",
  "data": {
    "OD": {
      "Circle Diameter": 3.5,
      "BMO Area": 1.82,
      "Classification": "Within Normal Limits",
      "Sector Map 1 Superior (S)": 105,
      "Sector Map 1 Nasal (N)": 57,
      "Sector Map 1 Temporal (T)": 57,
      "Sector Map 1 Inferior (I)": 109,
      "Sector Map 2 Nasal Superior (NS)": 98,
      "Sector Map 2 Temporal Superior (TS)": 115,
      "Sector Map 2 Nasal (N)": 63,
      "Sector Map 2 Global (G)": 82,
      "Sector Map 2 Temporal (T)": 57,
      "Sector Map 2 Nasal Inferior (NI)": 101,
      "Sector Map 2 Temporal Inferior (TI)": 123
    },
    "OS": {...}
  }
}
```

**Figure 3.** Schema-Constrained output example (RNFL Single Exam). Representative JSON showing valid, schema-compliant fields and values.

**Prompt-Only Parsing:** Extracted text was submitted to GPT-4o along with a detailed instruction prompt (Figure 4). The prompt specified the target fields and output conventions and returned loosely structured key-value text requiring post-processing (Figure 5).

```
system_prompt = """You are a helpful assistant extracting structured data from ophthalmology
reports."""
user_prompt = """, "RNFL Single Exam": """You are a medical assistant helping extract values from an
RNFL report. For each eye (OD, OS), extract: Circle Diameter; BMO Area;Classification; Sector Map 1: S,
 N, T, I; Sector Map 2: NS, TS, N, G, T, NI, TI. Output format: Eye:OD, ... ### Eye:OS, ...""",
```

**Figure 4.** Prompt-Only configuration. System and user prompts specifying target fields and formatting for key-value text output.

```
{
  "file_name": "redacted_122.pdf",
  "report_type": "RNFL Single Exam",
  "duration_s": 5.4,
  "parsed": "Eye:OD, Circle Diameter:3.5 mm, BMO Area:1.82 mm², Classification:Within Normal Limits,
Sector Map 1:S:105, N:57, T:57, I:109, Sector Map 2:NS:98, TS:115, N:63, G:82, T:57, NI:101, TI:123 ###
 Eye:OS, Circle Diameter:3.5 mm, BMO Area:1.62 mm², Classification:Outside Normal Limits, Sector Map
1:S:103, N:80, T:42, I:119, Sector Map 2:NS:119, TS:83, N:92, G:86, T:42, NI:111, TI:116"
}
```

**Figure 5.** Prompt-Only output example (RNFL Single Exam). Representative key-value string produced by the Prompt-Only method prior to rule-based conversion to JSON.

### Evaluation Metrics

The 20 reports were processed five times each with both parsing pipelines (prompt-only parsing and schema-constrained parsing). For any pipeline–report-type combination achieving less than 100% accuracy, discrepant fields were reviewed against the source PDF and ground truth. Each discrepancy was classified as a value error or a formatting error. Error frequency and affected fields were tabulated by report type and pipeline to identify systematic patterns.

**Value Accuracy (684 data points per trial):** The proportion of extracted values that matched the ground truth. Only the value component of each key-value pair was assessed.

**Formatting Accuracy (1,368 data points per trial):** The proportion of extracted values whose formatting (e.g., units, punctuation, field structure) matched the ground truth. Both the key and value components of each key-value pair were evaluated.

**Extraction Time (20 data points per trial):** The total time required for parsing each report.

All LLM processing was conducted using the GPT-4o model via the OpenAI API; no local installations or fine-tuning were performed.

## Results

**Dataset and runs.** Twenty de-identified diagnostic reports were analyzed (Visual Field, OCT Glaucoma Overview, OCT RNFL Single Exam, OCT Thickness Map; n = 5 each). Each report was processed five times per method (Prompt-Only and Schema-Constrained). Target fields are listed in Table 1.

**Value accuracy.** Initial manual transcription value accuracy was 98.32% for Visual Field, 99.13% for Glaucoma Overview, 95.17% for RNFL Single Exam, and 100.00% for Thickness Map. Schema-Constrained achieved 100.00% ± 0.00% value accuracy for Visual Field and OCT RNFL Single Exam reports. For Glaucoma Overview, Schema-Constrained achieved 97.45% ± 5.69% due to incorrectly extracted numeric values. For Thickness Map, Schema-Constrained achieved 98.00% ± 4.47% due to laterality confusion, in which OS values were assigned to the OD field and vice versa. Prompt-Only parsing achieved 100.00% ± 0.00% value accuracy across all report types (Table 2).

**Table 2.** Value accuracy (mean ± standard deviation, %) per report type by abstraction method: Manual, Schema-Constrained, and Prompt-Only. Abbreviations: OCT = optical coherence tomography; RNFL = retinal nerve fiber layer.

| Type of Report | Manual | Schema-Constrained | Prompt-Only |
|---|---|---|---|
| Visual Field | 98.32% | 100.00% ± 0.00% | 100.00% ± 0.00% |
| OCT Glaucoma Overview | 99.13% | 97.45% ± 5.69% | 100.00% ± 0.00% |
| OCT RNFL Single Exam | 95.17% | 100.00% ± 0.00% | 100.00% ± 0.00% |
| OCT Thickness Map | 100.00% | 98.00% ± 4.47% | 100.00% ± 0.00% |

**Formatting accuracy.** Manual transcription formatting accuracy was 99.16% for Visual Field and 100.00% for all other report types. Schema-Constrained parsing achieved 100.00% ± 0.00% formatting accuracy across all report types. Prompt-Only parsing achieved 100.00% ± 0.00% for Visual Field, Glaucoma Overview, and Thickness Map, and 90.14% ± 5.98% for RNFL Single Exam. The errors in RNFL reports came from omission of a required space between section headings and their values (e.g., "Sector Map:S:105" instead of the expected "Sector Map: S:105") (Table 3).

**Table 3.** Formatting accuracy (mean ± standard deviation, %) per report type by abstraction method: Manual, Schema-Constrained, and Prompt-Only. Abbreviations: OCT = optical coherence tomography; RNFL = retinal nerve fiber layer.

| Type of Report | Manual | Schema-Constrained | Prompt-Only |
|---|---|---|---|
| Visual Field | 99.16% | 100.00% ± 0.00% | 100.00% ± 0.00% |
| OCT Glaucoma Overview | 100.00% | 100.00% ± 0.00% | 100.00% ± 0.00% |
| OCT RNFL Single Exam | 100.00% | 100.00% ± 0.00% | 90.14% ± 5.98% |
| OCT Thickness Map | 100.00% | 100.00% ± 0.00% | 100.00% ± 0.00% |

**Extraction time.** Manual extraction times ranged from 11.09 s (Thickness Map) to 82.34 s (Visual Field), with a mean of 56.51 s per report. Schema-Constrained averaged 5.04 s per report (range 2.82–7.75 s), and Prompt-Only averaged 4.70 s (range 1.93–7.23 s), corresponding to mean reductions of ~92% vs. manual for both methods (Table 4).

**Table 4.** Mean extraction time (seconds ± standard deviation) per report by abstraction method: Manual, Schema-Constrained, and Prompt-Only. Abbreviations: OCT = optical coherence tomography; RNFL = retinal nerve fiber layer.

| Type of Report | Manual (s) | Schema-Constrained (s) | Prompt-Only (s) |
|---|---|---|---|
| Visual Field | 82.34 | 4.64 ± 1.08 | 7.23 ± 3.35 |
| OCT Glaucoma Overview | 64.27 | 4.94 ± 1.74 | 4.31 ± 0.87 |
| OCT RNFL Single Exam | 68.35 | 7.75 ± 2.56 | 5.31 ± 0.88 |
| OCT Thickness Map | 11.09 | 2.82 ± 0.89 | 1.93 ± 0.46 |

## Discussion

### Summary

This proof-of-concept study demonstrates that a general-purpose LLM can accurately and efficiently extract structured data from ophthalmic diagnostic reports, producing machine-readable outputs from PDF reports.

### LLM parsing may offer a practical alternative to manual abstraction

Due to incomplete adoption of data standards and challenges in integration and interoperability, discrete data from ophthalmic diagnostics are often not directly integrated into the EHR.[15] While manual transcription is sometimes used for clinical use, this process is labor-intensive, time-consuming, and prone to error, making it unsuitable for large datasets.[16] Studies have shown that human transcription error rates can exceed 20% even under double-review protocols, and manual extractors account for over a quarter of all discrepancies when compared with automated methods.[17,18]

In this study, automated parsing reduced mean per-report processing time by approximately 92% while maintaining high accuracy compared with the reconciled consensus ground truth. These findings suggest that general-purpose LLMs can achieve high accuracy in extracting structured values from semi-structured clinical reports in this pilot setting, offering a viable alternative to manual transcription. Comparable improvements have been observed in other domains: NLP systems for breast-cancer recurrence and medication abstraction have achieved equivalent or superior accuracy while reducing manual chart-review volume by nearly 90% and producing data consistent with manual abstraction for most inpatient medications.[17,19] These performance gains suggest potential operational benefits. General-purpose LLMs can process reports at scale, and in doing so shift human effort from transcription toward targeted validation and quality control.[16,18] This scalability supports rapid dataset creation for both retrospective analyses and prospective studies, facilitating faster research cycles and continuous dataset updates with limited staff involvement.[19]

### Performance was consistent across the four report formats tested, suggesting that the approach may be adaptable to additional report types with limited engineering

The LLM system evaluated in this study achieved high performance across four distinct diagnostic report types (Visual Field, OCT Glaucoma Overview, OCT RNFL Single Exam, and OCT Thickness Map) demonstrating 100% formatting accuracy for Schema-Constrained across all four report types and 100% formatting accuracy for Prompt-Only on three report types (90.14% for RNFL Single Exam) without model fine-tuning or architecture changes. While the target fields for extraction were tailored to each report format, the overall system design, including prompt structure and parsing strategy, remained unchanged. This adaptability suggests that similar pipelines may be deployable across heterogeneous imaging devices or evolving software versions with limited development overhead. Adapting the pipeline to new report formats could require relatively minor modifications rather than extensive system redevelopment, supporting scalable implementation in both research and clinical settings.

This portability has practical implications for scaling structured data extraction pipelines in real-world ophthalmology environments, where imaging devices and software versions can vary across and within institutions. Systems that require minimal engineering effort to accommodate new formats reduce implementation barriers and speed up deployment timelines.[20] Moreover, adaptability across distinct input styles may mitigate the brittleness often associated with deterministic rule-based approaches, which can fail when faced with layout or phrasing variation.[7,8,21] These findings complement prior work in other clinical domains that also underscore the potential of general-purpose LLMs to perform robustly across heterogeneous clinical document types without retraining.[10,11,19]

**This approach may facilitate standardization and interoperability by converting heterogeneous report content into structured, machine-readable data**

The structured outputs produced by this pipeline also point toward broader goals of data standardization and interoperability in ophthalmology. By converting human-readable diagnostic PDFs into machine-readable formats (e.g., JSON or key–value representations), systems like the one evaluated here could, in principle, facilitate data exchange among EHRs, registries, research databases, and other downstream systems.[4–7]

Because these outputs are structured and timestamped, they could support auditability and reproducibility and may facilitate compliance with applicable regulatory requirements.[22,23] In clinical practice, this could enable automated population of structured fields, longitudinal tracking of disease metrics, and integration of extracted diagnostic data into downstream decision-support systems.[8,23] Because the system can parse multiple report formats and consistently output structured representations, these outputs could serve as an intermediate step toward mapping vendor-specific report data to interoperability standards such as DICOM attributes or HL7 FHIR resources.[7]

These same properties can also benefit research. Retrospective studies often rely on harmonizing large volumes of imaging data generated across devices, software versions, and time periods. Structured extraction could enable aggregation of heterogeneous data into analyzable cohorts—supporting retrospective registry construction, multisite research collaboration, and large-scale quality improvement initiatives.[4,23,24]

Taken together, these capabilities lay the groundwork for more connected, scalable, and standards-aligned ophthalmic data infrastructure across both clinical and research domains.

### Limitations and Future Directions

This initial pilot study used a small sample (n = 20) of de-identified reports intentionally selected for embedded machine-readable text; scanned image PDFs and handwritten annotations were excluded. Only one model (GPT-4o) and two prompting strategies were evaluated. Additionally, this study evaluated four report types at one institution, and performance may differ across models, institutions, vendors, and report formats. Despite evaluating only GPT-4o, we observed consistently high accuracy across the tested report types. Future work should investigate whether these findings extend to newer models.

These constraints limit generalizability but also demonstrate feasibility with minimal configuration. Future work will expand evaluation to larger and more diverse datasets, incorporate OCR and multimodal pipelines for image-only reports, and compare alternative models on accuracy, latency, and cost. Prospective testing in real-world environments will further assess robustness and the potential of LLM-assisted parsing to streamline ophthalmic data integration at scale.

In this pilot evaluation, LLM-assisted parsing extracted discrete data from human-readable ophthalmic diagnostic PDFs with high accuracy and substantially reduced processing time. These findings support further investigation of schema-guided extraction and targeted validation as potential steps toward closing the long-standing interoperability gap in ophthalmic research and patient care.

## Funding

Supported by the National Institutes of Health (NIH), Bethesda, MD (R01EY030929, R01EY013178, R50EY038047, R21EY037312, T15LM007088, P30EY010572 [OHSU core grant], and P30EY001792 [UIC core grant]); the Malcolm M. Marquis, MD Endowed Fund for Innovation; and unrestricted departmental grants from Research to Prevent Blindness, New York, NY, to the Casey Eye Institute, Department of Ophthalmology,

Oregon Health & Science University, and to the Department of Ophthalmology and Visual Sciences, University of Illinois Chicago. The sponsors and funding organizations had no role in the design or conduct of this research.

## Conflict of Interest

No conflicting relationship exists for any author.

## Ethics Approval

This study adhered to the tenets of the Declaration of Helsinki and was approved by the Institutional Review Board at Oregon Health & Science University (STUDY00026815) and by the AI Governance Committee.

## Abbreviations

AI = artificial intelligence; API = application programming interface; DICOM = Digital Imaging and Communications in Medicine; EHR = electronic health record; GHT = Glaucoma Hemifield Test; HAAT = Horizontal Angle Asymmetry Test; HIPAA = Health Insurance Portability and Accountability Act; HL7-FHIR = Health Level Seven Fast Healthcare Interoperability Resources; IRB = Institutional Review Board; JSON = JavaScript Object Notation; LLM = large language model; MD = Mean Deviation; MRW = Minimum Rim Width; NLP = natural language processing; OCR = optical character recognition; OCT = optical coherence tomography; OD = oculus dexter; OHSU = Oregon Health & Science University; OS = oculus sinister; PDF = portable document format; PSD = Pattern Standard Deviation; RNFL = retinal nerve fiber layer; RNFLT = retinal nerve fiber layer thickness; VFI = Visual Field Index